\documentclass[nonacm,dvipsnames,format=sigconf,anonymous=false,review=false]{acmart}
\AtBeginDocument{%
  }

\setcopyright{acmlicensed}
\copyrightyear{2025}
\acmYear{2025}
\setcopyright{cc}
\setcctype{by}
\acmConference[GECCO '25]{Genetic and Evolutionary Computation
Conference}{July 14--18, 2025}{Malaga, Spain}
\acmBooktitle{Genetic and Evolutionary Computation Conference (GECCO '25),
July 14--18, 2025, Malaga, Spain}
\acmDOI{10.1145/3712256.3726436}
\acmISBN{979-8-4007-1465-8/2025/07}

\usepackage[noabbrev, capitalise]{cleveref}

\newcommand{\gab}{$G_{AB}$}

\begin{document}


\title{To Repair or Not to Repair? Investigating the Importance of AB-Cycles for the State-of-the-Art TSP Heuristic EAX}


\author{Jonathan Heins}
\orcid{0000-0002-3571-667X}
\affiliation{%
  \institution{"Friedrich List" Faculty of Transport and Traffic Sciences, TU Dresden}
  \city{Dresden}
  \country{Germany}
}
\email{jonathan.heins@tu-dresden.de}

\author{Darrell Whitley}
\affiliation{%
  \institution{Department of Computer Science, Colorado State University}
  \city{Fort Collins}
  \state{CO}
  \country{USA}}
\email{whitley@cs.colostate.edu}

\author{Pascal Kerschke}
\affiliation{%
  \institution{"Friedrich List" Faculty of Transport and Traffic Sciences, TU Dresden \& ScaDS.AI Dresden/Leipzig}
  \city{Dresden}
  \country{Germany}
}
\email{pascal.kerschke@tu-dresden.de}







\begin{abstract}

The Edge Assembly Crossover (EAX) algorithm is the state-of-the-art heuristic for solving the Traveling Salesperson Problem (TSP). It regularly outperforms other methods, such as the Lin-Kernighan-Helsgaun heuristic (LKH), across diverse sets of TSP instances. Essentially, EAX employs a two-stage mechanism that focuses on improving the current solutions, first, at the local and, subsequently, at the global level. Although the second phase of the algorithm has been thoroughly studied, configured, and refined in the past, in particular, its first stage has hardly been examined. 

In this paper, we thus focus on the first stage of EAX and introduce a novel method that quickly verifies whether the AB-cycles, generated during its internal optimization procedure, yield valid tours -- or whether they need to be repaired. Knowledge of the latter is also particularly relevant before applying other powerful crossover operators such as the Generalized Partition Crossover (GPX). Based on our insights, we propose and evaluate several improved versions of EAX.
According to our benchmark study across $10\,000$ different TSP instances, the most promising of our proposed EAX variants demonstrates improved computational efficiency and solution quality on previously rather difficult instances compared to the current state-of-the-art EAX algorithm.
\end{abstract}




\maketitle

\section{Introduction}

The \textit{Traveling Salesperson Problem (TSP)} is a classical NP-hard combinatorial optimization problem that is relevant in various fields, e.g., in computer science, logistics, and transportation science \cite{applegate2011traveling}. Apart from the manifold of application areas, the beauty of this problem lies in the simplicity of its underlying task. Let $V := \{v_1, \ldots, v_N\}$ be a set of $N$ nodes (or vertices), $E := \{e_k = (v_i, v_j) \in V \times V\}$ a set of edges connecting all pairwise combinations of the nodes, and $c: E \rightarrow \mathbb{R}$ a cost function that quantifies the cost (e.g., distance or duration) of traveling along an edge. Then, the goal of the TSP is to find a round trip along all nodes -- i.e., a path or sequence of adjacent edges where the first and the last node of the sequence are identical -- so that each node is visited exactly once and the cost for the entire tour is minimal.

In recent years, research on TSP has mainly focused on two perspectives. A few publications focused on a better understanding of the problem structures and how they can reveal strengths and weaknesses of search algorithms \cite{heins2023, bossek2019}. In contrast, the majority of works focused on the TSP solvers themselves, either by analyzing state-of-the-art solvers and refining parts of them, or by proposing new algorithmic approaches \cite{heins2024dancing, min2024unsupervised, varadarajan2022mixing, whitley2022local, mukhopadhyay2021efficient,tinos2020new,varadarajan2020many,varadarajan2019}. Despite the wealth of studies dealing with the latter perspective, the state of the art in heuristic TSP solving can be reduced to two algorithms: the \textit{Edge Assembly Crossover (EAX)} algorithm \cite{nagata1997edge,nagata2013powerful} and the \textit{Lin-Kernighan-Helsgaun (LKH)} heuristic \cite{helsgaun2000effective,helsgaun2009general}. Although EAX initially denoted only the crossover operation, the term is now used for the full genetic algorithm -- a convention that we adopt throughout this paper. In particular, restart versions of these two solvers \cite{duboislacoste2015} showed superior performance in various benchmark studies covering thousands of instances from different TSP sets \cite{kotthoff2015improving,kerschke2018leveraging,heins2023}.

A very recent paper offered new insights into the inner mechanics of LKH \cite{heins2024dancing}, helping to make it more competitive with EAX -- especially on instances constructed in favor of EAX. Still, despite these improvements, EAX remains the single-best TSP heuristic. 
And while multiple works studied and improved LKH in recent years \cite{heins2024dancing,mukhopadhyay2021efficient,taillard2019popmusic}, many building blocks of EAX have hardly been analyzed, let alone optimized in a targeted manner. With this work, we aim to address this research gap by thoroughly investigating the first stage of EAX. We then use our findings to propose a competitive EAX variant that specifically leverages EAX performance on TSP instances with 500 nodes and on instances that were rather favorable for LKH.

The remainder of this work is structured as follows. \cref{sec:background} introduces relevant concepts of solving the TSP using crossover operators, and summarizes the corresponding background. EAX itself is then outlined in \cref{sec:stages}. In \cref{sec:examination}, we analyze stage I of EAX in more detail using an experimental study. Subsequently, we introduce the concept for a fast check of valid tours in \cref{sec:check}, which enables the implementation of faster EAX alternatives. Based on our findings, we propose and benchmark alternative EAX variants in \cref{sec:variants} and conclude our work in \cref{sec:conclusion}.

\section{Background}
\label{sec:background}

\begin{figure}[t]
    \centering
    \includegraphics[width=0.99\columnwidth]{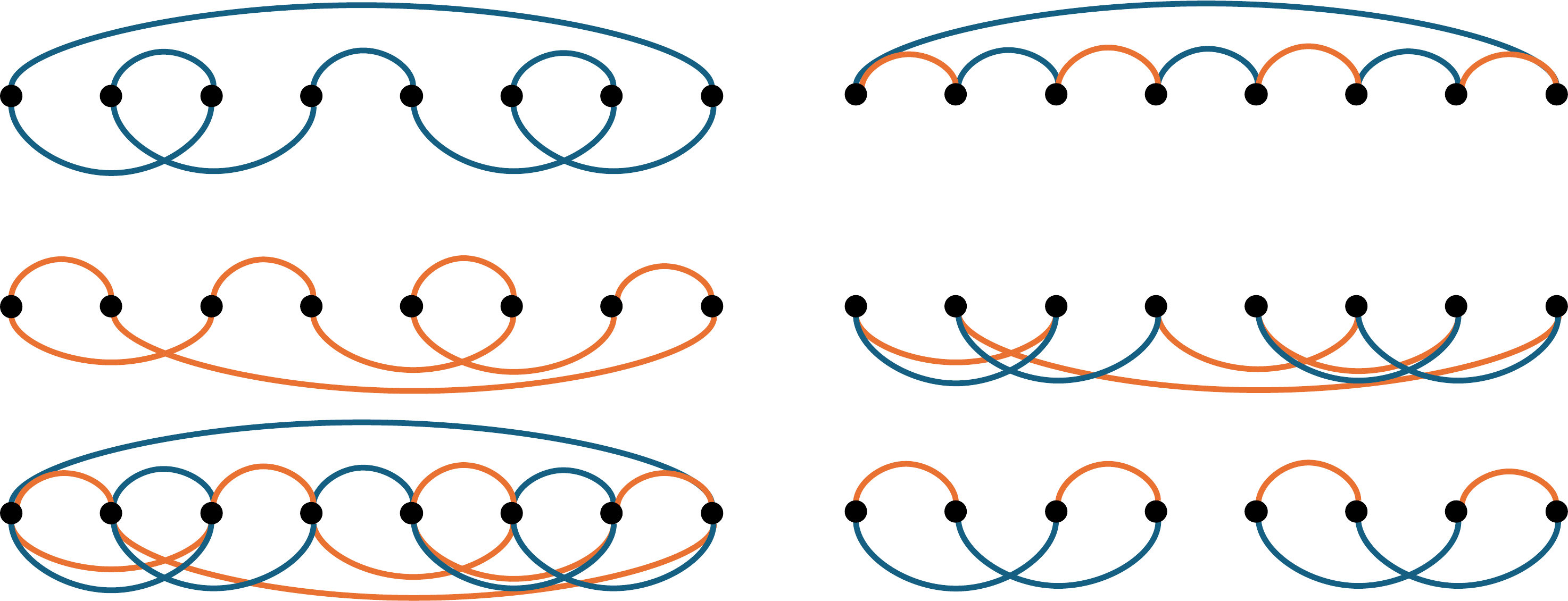}
    \caption{Left: Parents $A$ (blue) and $B$ (orange) form \gab, which contains all edges (bottom). Right: The AB-cycle (top), the (simplified) outer graphs (middle), and the invalid solution that would result from applying the AB-cycle to $A$.}
    \label{fig:ab-cycle-mirrorex}
\end{figure}

Most genetic algorithms for solving the TSP rely on powerful crossover operators -- such as IPT \cite{mukhopadhyay2021efficient}, GPX \cite{gpx1, tinos2020new}, and EAX \cite{nagata1997edge,nagata2013powerful} -- which efficiently process so-called AB-cycles.
As a result, they have been 
examined 
in many previous works \cite{ozeas2019, nagata1997edge, nagata2013powerful, gpx1, tinos2020new}.

Let $E_A$ and $E_B$ be the edge sets of two (parent) solutions $A$ and $B$, respectively, for a given TSP instance. 
The corresponding union graph $G_{AB} = (V, E_A \cup E_B)$ contains all distinct edges from both parents. 
See the left side of \cref{fig:ab-cycle-mirrorex} for an example of $A$, $B$ and the resulting graph \gab.
Then, an \textit{AB-cycle} is any cycle in \gab\ with alternating edges from $E_A$ and $E_B$, see, e.g., the top right image of \cref{fig:ab-cycle-mirrorex}.
These AB-cycles are typically applied to one or both parent solutions to produce offspring.
That is, if the AB-cycle $c_{AB}$ (with edges $E_{c_{AB}}$) is combined with the parent solution $A$, a new (offspring) solution $C$ is obtained. The edge set $E_C$ of $C$ is based on the edges from $A$, except for edges, which were also contained in the AB-cycle $c_{AB}$. The `missing' edges are replaced by the remaining edges of the AB-cycle, which were caused by $B$. Hence, $C$ consists of the edges $E_C := (E_A \setminus (E_A \cap E_{c_{AB}})) \cup (E_B \cap E_{c_{AB}})$.
However, the resulting solution is not guaranteed to be feasible as the exchange of edges may produce two or more disconnected subtours, thereby violating the requirement for a Hamiltonian cycle, as depicted, e.g., in the bottom right image of \cref{fig:ab-cycle-mirrorex}. 

\begin{figure}[t]
    \centering
    \includegraphics[width=0.8\columnwidth]{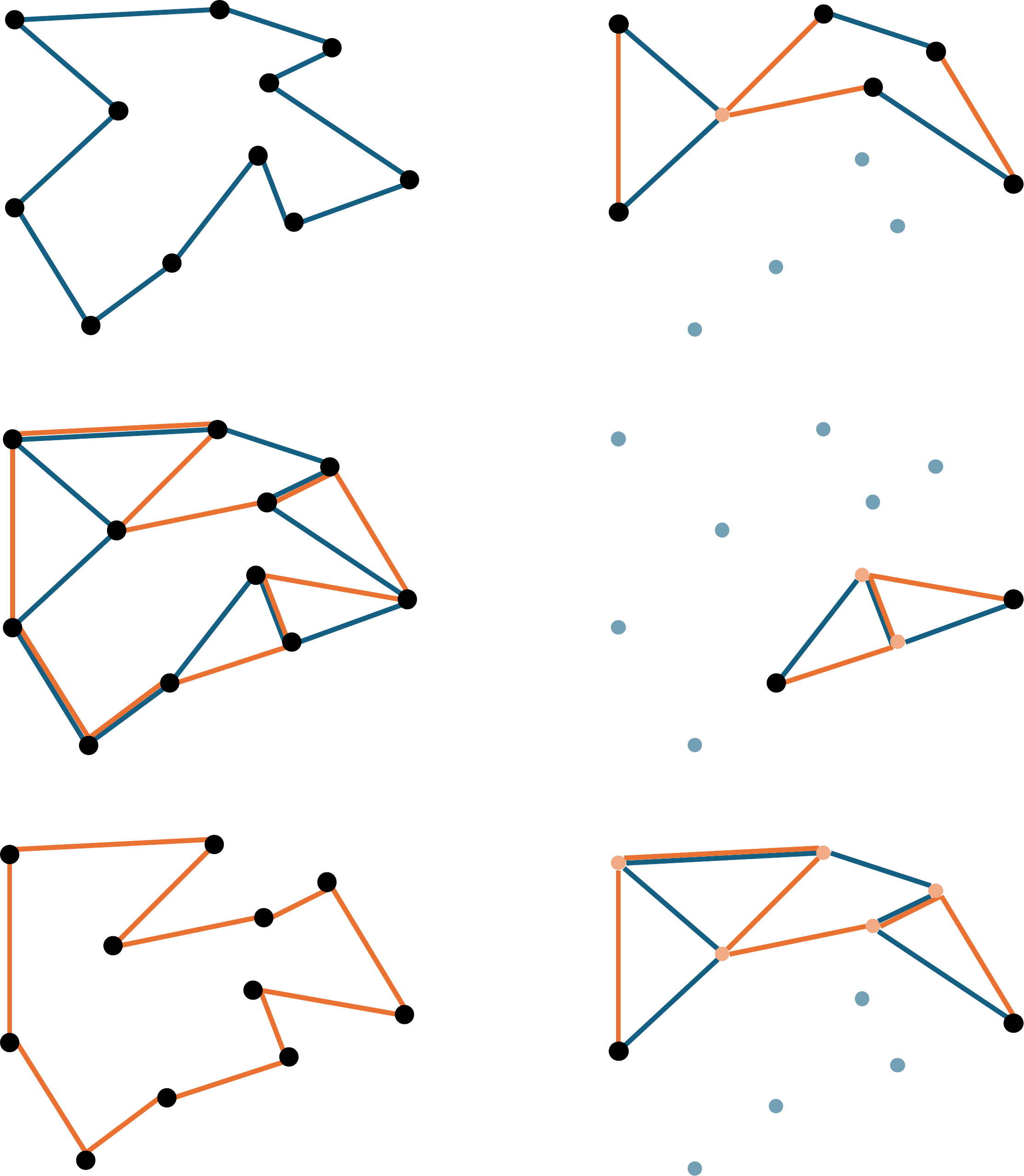}
    \caption{Left: Parent $A$ (blue, top) and Parent $B$ (orange, bottom) form \gab, which contains all edges (middle). Right: Examples of AB-cycles resulting in different E-sets used by EAX. Points are distinguished into $A$- (blue), $B$- (red), and portals / $C$-vertices (black) in the corresponding E-set.}
    \label{fig:eset}
\end{figure}

The EAX algorithm employs a repair mechanism to transform such invalid solutions into feasible ones. In contrast, IPT and GPX restrict AB-cycles to those that directly result in Hamiltonian cycles, requiring a feasibility check before application. 
Such checks can be facilitated by making use of the concept of internal and external edges, as well as portals of AB-cycles. 
Internal edges are edges that are included in $E_{c_{AB}}$, whereas external edges are not included in $E_{c_{AB}}$. Portals are vertices that serve as transitions between external and internal edges; specifically, they are connected to two external edges and two internal edges. In general, every vertex in \gab\ is connected to zero, two, or four internal edges. This is caused by the fact that an AB-cycle either does not pass through a vertex, or enters and exits it exactly once or twice 
via alternating edges from $E_A$ and $E_B$, see \cref{fig:eset}. If there are exactly two portals -- see, e.g., the middle and bottom image on the right of \cref{fig:eset} -- the resulting solution forms a complete tour as, in this case, the application of the AB-cycle merely rearranges the order of internal vertices between the two portal nodes. 
Hence, there are only two loose ends of the overall circle formed by parent $A$, and reconnecting them with an altered path cannot break the connectiveness, as the same vertices are visited just in a different order.
This observation is utilized in the original versions of GPX and IPT to guarantee the production of valid offspring. 
In the second version of GPX, AB-cycles with more portals -- see, e.g., image on the top right of \cref{fig:eset} -- are allowed, if the simplified internal graph is the same. That is, the undirected links formed by internal edges between the portals are the same in $A$ and $B$. In such cases, the original tour is cut in a way that could potentially form subtours; however, since the same endpoints are reconnected, the resulting tour remains feasible. 

Having only two portals in an AB-cycle, or identical simplified internal graphs are two sufficient conditions to ensure a valid tour. However, neither is a necessary condition \cite{ozeas2019}. \citeauthor{ozeas2019} proposed an alternative approach, known as the mirror-test, which leverages the necessary condition that no two links can be identical between the internal graph of one parent and the external graph of the other parent (i.e., the undirected links between portals formed by external edges). Although this condition is necessary, it is not sufficient for all possible pairs of parent tours. The original paper does not explicitly acknowledge this limitation; however, a counterexample is provided in \cref{fig:ab-cycle-mirrorex}. 

In this paper, we introduce a test that is both necessary and sufficient to ensure that applying an AB-cycle results in a valid tour. While EAX includes a repair mechanism to address invalid tours that are broken into subtours, it avoids reliance on such tests. However, as repairs are computationally expensive, EAX employs additional strategies, detailed in the next section, to minimize the expected number of subtours.

\section{The two Stages of EAX}
\label{sec:stages}

\begin{figure}[t]
    \centering
    \includegraphics[width=0.85\columnwidth]{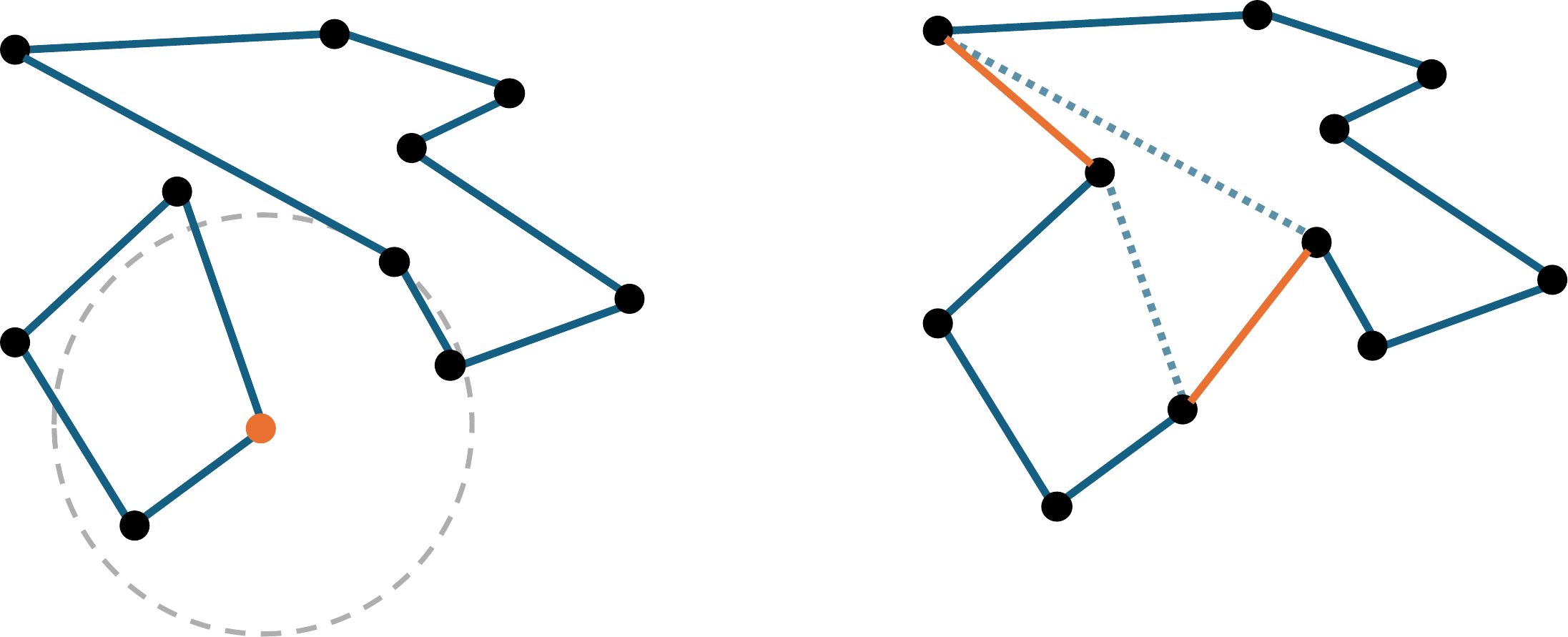}
    \caption{Left: If the application of an AB-cycle leads to two or more subtours the solution is repaired by looking at the ten nearest neighbors of every node of the smallest subtour (here depicted by the dashed circle at the 4th nearest neighbor distance of the orange node) to find the most beneficial tuple of four nodes that remove two edges and add two edges to connect two subtours (right).}
    \label{fig:repair}
\end{figure}

The EAX algorithm is based on two stages.
The first stage (see~Section~\ref{sec:stagei}) focuses on improving the solutions locally, while the second stage (see~Section~\ref{sec:stageii}) aims at global optimization. 
The key distinction between these stages lies in the construction of the set of applied AB-cycles.
This section provides a brief overview of the most important concepts discussed in this paper.
The parameters introduced in the section are the default settings proposed by \citeauthor{nagata2013powerful}.
For further details, interested readers are referred to the original publication \cite{nagata2013powerful,nagata1997edge}.

In a nutshell, EAX searches for an AB-cycle and applies it to parent $A$.
\gab\ can potentially be split into multiple
configurations of AB-cycles and the configuration used by EAX is generated by randomly tracing a path of alternating edges. Once a cycle is detected, it is identified as an AB-cycle.
After adding this cycle to the set of AB-cycles, the tracing process continues from the last vertex not included in the current AB-cycle, or from a randomly selected new starting vertex.
The E-set is the set of AB-cycles that is applied to parent $A$ -- see \cref{fig:eset} for an exemplary illustration of E-sets.
The construction of this set, thus, determines the strategy of EAX. 

\subsection{Stage I}
\label{sec:stagei}

In stage one, only one randomly chosen AB-cycle is applied to parent $A$ to generate one offspring.
This is argued to be a local optimization as the resulting offspring tends to have many edges of $A$ and only few of $B$.
If the Application of the AB-cycle results in two or more subtours, the algorithm repairs the solution by iteratively reconnecting the different subtours with each other. The procedure starts with the smallest subtour and searches for one edge to be deleted along with a close edge of another subtour. The four vertices involved in that deletion procedure are then reconnected with two new edges that unify the two subtours into one. The search aims to finding four such edges -- two to be added and two to be removed --
that result in the shortest tour length of the offspring solution. 

As an extensive search would be too costly, the edge to be removed from the neighboring subtour is required to connect to at least one vertex in the 10th nearest neighbor-neighborhood of one of the two vertices of the edge to be removed in the smallest subtour. That is, all ten nearest neighbors
of all vertices in the smallest subtour are searched as a starting point of a candidate edge. This procedure is sketched in \cref{fig:repair} (using the 4th nearest neighbor-neighborhood for better readability).

Once the two subtours are merged the now smallest subtour is merged with another subtour. This procedure will be repeated until the complete solution is a Hamiltonian cycle. Depending on the number of subtours this can be an expensive operation and does not necessarily yield good new edges. Hence, this approach is avoided in the global search stage.

\subsection{Stage II}
\label{sec:stageii}

In stage two of EAX, multiple AB-cycles are chosen and combined to create an AB-cycle that introduces as few subtours as possible. To find an appropriate E-set, the number of so-called $A$-, $B$-, and $C$-vertices are examined to determine how many subtours could potentially result from applying the corresponding AB-cycles \cite{nagata2013powerful}. 

All nodes that are not in the E-set are an $A$-vertex (external node), all nodes with four links in the E-set are a $B$-vertex (internal node), and all nodes with two links are a $C$-vertex, as depicted in \cref{fig:eset}.  Note that the $C$-vertices correspond to the portals from the GPX-notation. The number of $C$-vertices $\#C$ determines the maximal number of resulting subtours, denoted as  $m$, where $m \leq \#C / 2$ \cite{nagata2013powerful}. Therefore, in stage two, EAX attempts to combine AB-cycles that share $C$-vertices to reduce $\#C$ down to $2$, which would result in a single subtour. To achieve this, a tabu search is employed, to maintain a feasible search space. If no valid combination for $\#C = 2$ is identified after 20 iterations, the search terminates with the configuration with minimal $\#C$. 

\section{Examining Stage I of EAX}
\label{sec:examination}

Stage I of EAX serves as the local optimization phase, as selecting only a single AB-cycle to form the complete E-set typically results in a new individual that retains many edges from parent $A$ while incorporating relatively few from parent $B$ \cite{nagata2013powerful}. However, the effects of different AB-cycle types have not been previously explored.

\begin{figure*}[t]
    \centering
    \includegraphics[width=\linewidth]{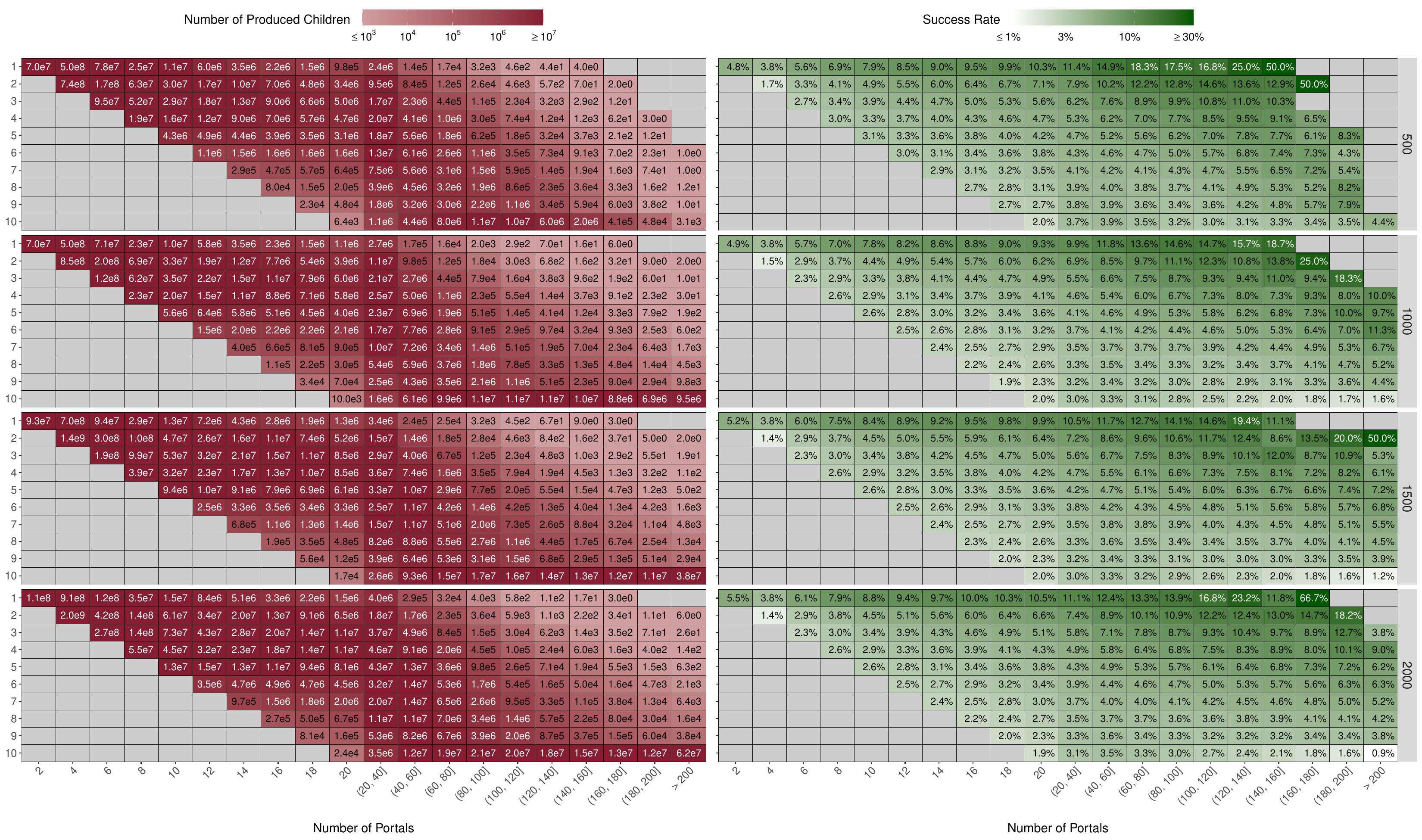}
    \caption{Left: The total number of all children produced in stage I of EAX, differentiated by the AB-cycle type producing the offspring and distinguished by the instance sizes. On the $x$-axis are the number of portals or $C$-vertices in the AB-cycle and on the $y$-axis the number of subtours, which had to be joined to produce a valid solution. Right: The percentage of all offspring from the left side, which were selected to replace their respective parent $A$ (success rate). Interestingly, the AB-cycle type that is found most often with four portals and two subtours has one of the lowest success rates.}
    \label{fig:heatmap}
\end{figure*}

In this section, we analyze AB-cycles based on the number of portals (or $C$-vertices) and the number of subtours produced when the AB-cycle is applied to parent $A$. The goal is to quickly identify unpromising AB-cycles that can be disregarded and replaced with more advantageous alternatives, thereby improving the efficiency of the recombination process.

For this purpose, we conducted an experimental study, in which we executed the original EAX on a diverse set of $10\,000$ evolved TSP instances. The underlying instance generation method was introduced in \cite{bossek2019}. Based on this method, \cite{heins2021} created the set of $10\,000$ instances used in this work. Those instances were optimized using random uniform mutations (simple) or advanced operations such as explosion or implosion (sophisticated) to be either difficult to solve for LKH and easy for EAX (EAX-friendly instances) and vice versa (LKH-friendly instances). Thus, the instances are well suited to study algorithm behavior as they contain both very easy and challenging scenarios. The evolution process is repeated for four different instance sizes ($500,\ 1\,000,\ 1\,500,\ 2\,000$) and every one of those size evolution type combination is represented by $500$ instances except for all groups of size $500$ which are represented by $1\,000$ instances.  
Across all the runs, we then recorded the total number of AB-cycles produced per type of AB-cycle (defined by the combination of the number of portals and subtours), and the percentage of those offspring that were selected to replace its parent. The corresponding results, distinguished by the size of the underlying TSP instances, are depicted in \cref{fig:heatmap}.
Due to the constraint $m \leq \#C / 2 $ described in \cref{sec:stageii}, no AB-cycles that produce more subtours than half of the number of portals can be recorded. 
Additionally, larger problem instances tend to exhibit larger AB-cycles in terms of portals, though these cycles are less likely to yield valid solutions. However, when a valid solution is achieved, it has a higher probability of outperforming offspring generated by smaller AB-cycles.

\begin{figure*}[t]
    \centering
    \includegraphics[width=\linewidth]{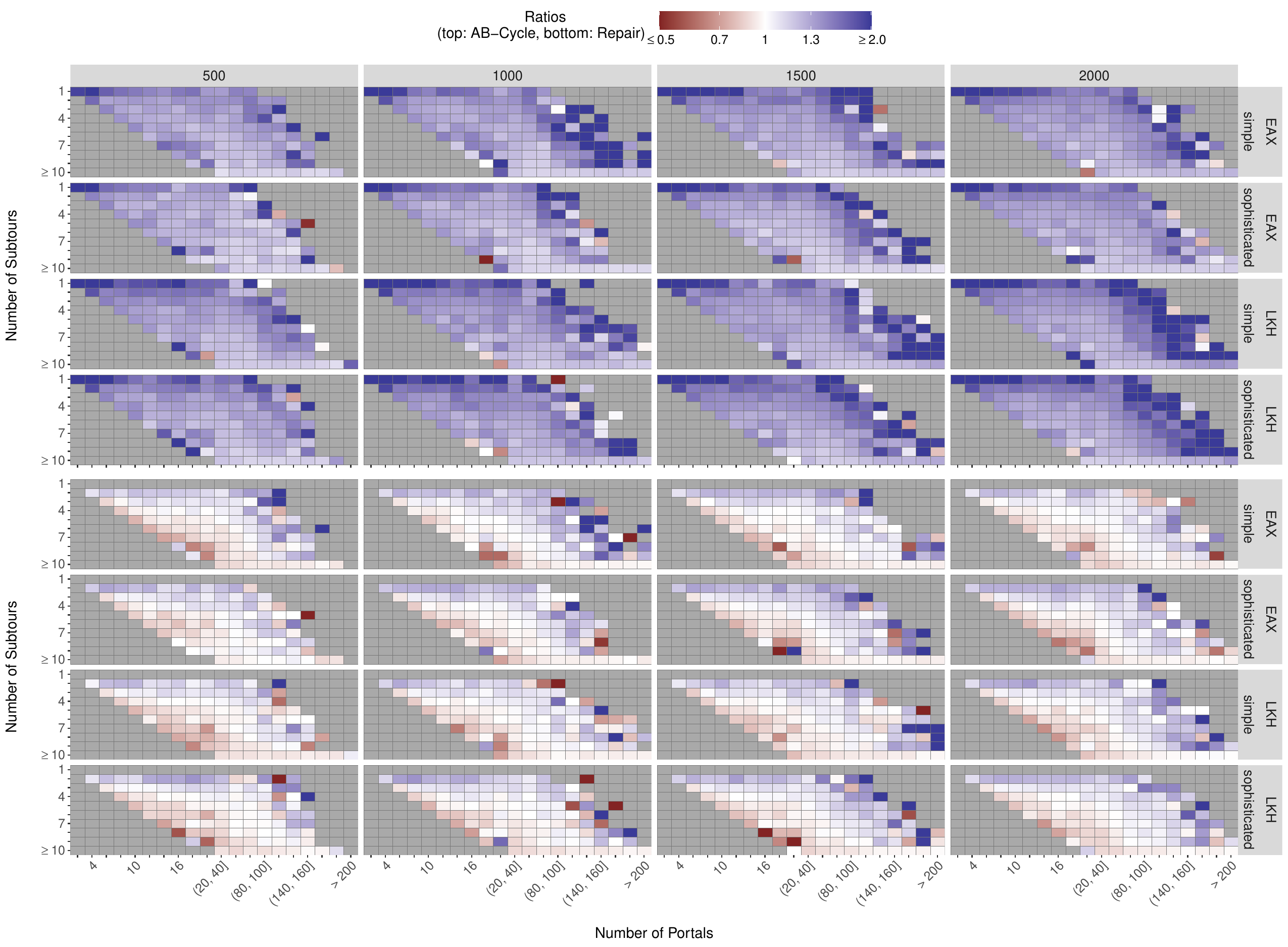}
    \caption{Number of optimal edges gained vs. lost distinguished by the size (columns) and type (rows) of TSP set, as well as the number of portals (x-axis) and produced subtours (y-axis) per AB-cycle. Top: Ratio of all optimal edges introduced (vs. lost) by the specific AB-cycle type, ignoring edges that are removed or added in a potential repair mechanism. Bottom: Average ratio of all optimal edges introduced (vs. lost) during the corresponding repair mechanism. Notably, AB-cycle types that introduce relatively few subtours compared to the number of subtours that could potentially be generated appear to be favorable.}
    \label{fig:heatmap_ratios}
\end{figure*}

The success rate itself does not necessarily indicate that the selected AB-cycle contributes to reaching the global optimum. It is possible that EAX fails to find the shortest Hamiltonian cycle because it frequently selects an unhelpful AB-cycle type. To address this, \cref{fig:heatmap_ratios} illustrates the effectiveness of each AB-cycle type. It depicts how many optimal edges -- absent in parent $A$ -- are introduced into the offspring and how many are lost.
Optimal edges can be gained or lost in two ways: either they are part of the AB-cycle itself, or they are modified during the repair mechanism. An AB-cycle improves an individual's proximity to the optimal solution, if it introduces more optimal edges than it removes. Thus, AB-cycles can be evaluated by computing the ratio of gained optimal edges to lost optimal edges -- separately for those within the AB-cycle (upper half of \cref{fig:heatmap_ratios}) and those affected by the repair process (lower half).
As shown in \Cref{fig:heatmap_ratios}, this ratio strongly depends on the AB-cycle type. AB-cycles that produce a complete tour -- corresponding to a single subtour (see top row per heatmap) -- generally introduce a higher number of optimal edges, and also to those that require minimal repairs relative to the number of portals. Notably, this is true for the number of edges introduced with the AB-cycle and with the repair mechanism. In contrast, AB-cycles that generate the maximum possible number of subtours given the number of portals tend to introduce only a few additional optimal edges while losing more optimal edges through the repair mechanism (indicated by reddish tiles).

\section{A Novel Check for Valid Tours}
\label{sec:check}

So far, the justification for using $C$-vertices to obtain an upper limit for the number of generated subtours is based on the fact that it would be too time-consuming to check whether an AB-cycle leads to a valid tour (or not) before applying it.
However, as our experiments show with a fast implementation of a checking procedure, the code requires less time for checking whether an AB-cycle would result in a complete tour compared to the time it takes for repairing a non-valid tour. In this paper, we only apply this check to the first stage of EAX as most repairs are necessary because of single AB-cycles. However, a new search guided by our checking procedure, replacing tabu search in stage II, is a possible extension of our work and another research direction.
Our check relies on the fact that it is sufficient to test only the $C$-vertices of an AB-cycle to determine whether its application will produce a valid tour.
\textbf{Theorem} If an AB-cycle $c_{AB}$, produced by tours $A$ and $B$, is applied to tour $A$, it is \textit{necessary} and \textit{sufficient} to only examine whether the simplified outer connections of $A$ joined with the simplified inner connections of $B$ (based on a set of $C$-vertices) form a Hamiltonian circuit.

\textbf{Proof} 
Let tours $A$ and $B$ be represented by the sequence of vertices $(a_1,\dots, a_N)$ and $(b_1, \ldots, b_N)$, respectively, where $N$ is the number of nodes. 
Further, let the AB-cycle $c_{AB}$ be the set of paths 
\begin{eqnarray*}
   \{(a_{c_1},\dots, a_{c_2}),\dots,(a_{c_{n-1}},\dots, a_{c_n}),\\
   \phantom{\{}(b_{c_1},\dots b_{c_2}),\dots,(b_{c_{n-1}},\dots, b_{c_n})\},
\end{eqnarray*}
where $c_{AB}$ contains $n$ $C$-vertices, and these paths represent segments of $A$ and $B$ over the same set of nodes and every path starts in a portal and ends in another.

\begin{figure}[t]
    \centering
    \includegraphics[width=0.8\columnwidth]{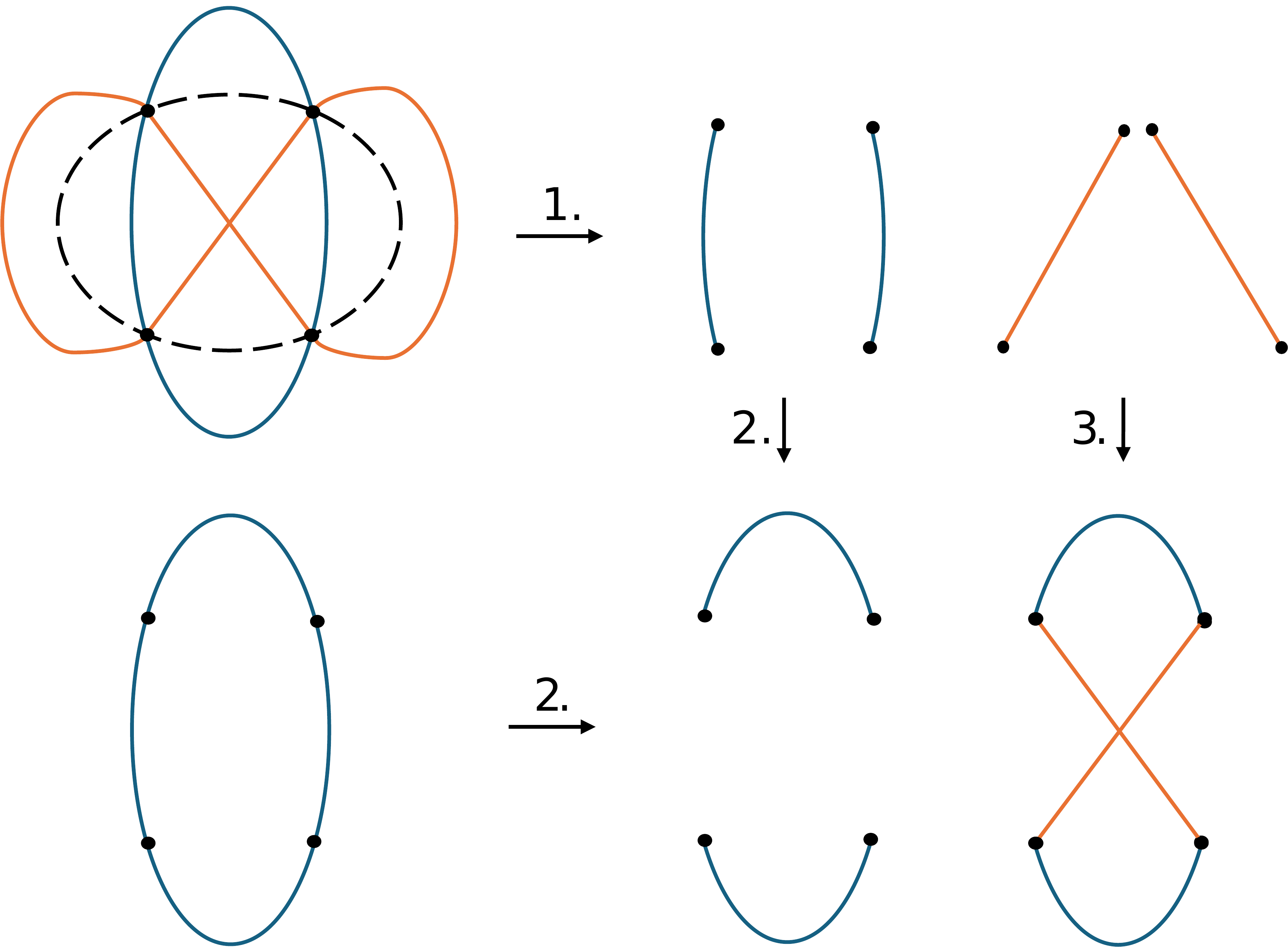}
    \caption{
    Top: An AB-cycle $c_{AB}$ is the result of cutting \gab\ at the portal vertices (dashed line through black dots), producing a set of paths from $A$ (blue) and $B$ (orange), each connecting two portals (step 1). Bottom: Starting with tour $A$ (left), applying the AB-cycle involves first removing the paths in $c_{AB}$ that originated from $A$ (step 2), leaving unconnected paths that still link two portals each (center). Finally, in step 3, the unconnected paths from $A$ are joined with the corresponding $B$-paths from $c_{AB}$ by connecting them at matching portal vertices, forming the final structure (right).
    }
    \label{fig:abcycle_paths}
\end{figure}

Applying $c_{AB}$ to $A$ as depicted in \cref{fig:abcycle_paths} involves the following steps: First, remove from $A$ the paths from $c_{AB}$, resulting in a set of disjoint paths $\{(a_1,\dots, a_{c_1}),\dots,(a_{c_{n}},\dots, a_N)\}$.
Second, integrate the $B$-paths from $c_{AB}$ into the set of remaining paths from $A$. 
The paths are joined by matching endpoints. Thus, for two paths $(a_{c_l},\dots,a_{c_i})$ and $(b_{c_j},\dots,b_{c_k})$, where $a_{c_i} = b_{c_j}$, the resulting path will be $(a_{c_l},\dots, b_{c_j}, \dots,b_{c_k})$.

\begin{figure}[t]
    \centering
    \includegraphics[width=0.75\columnwidth]{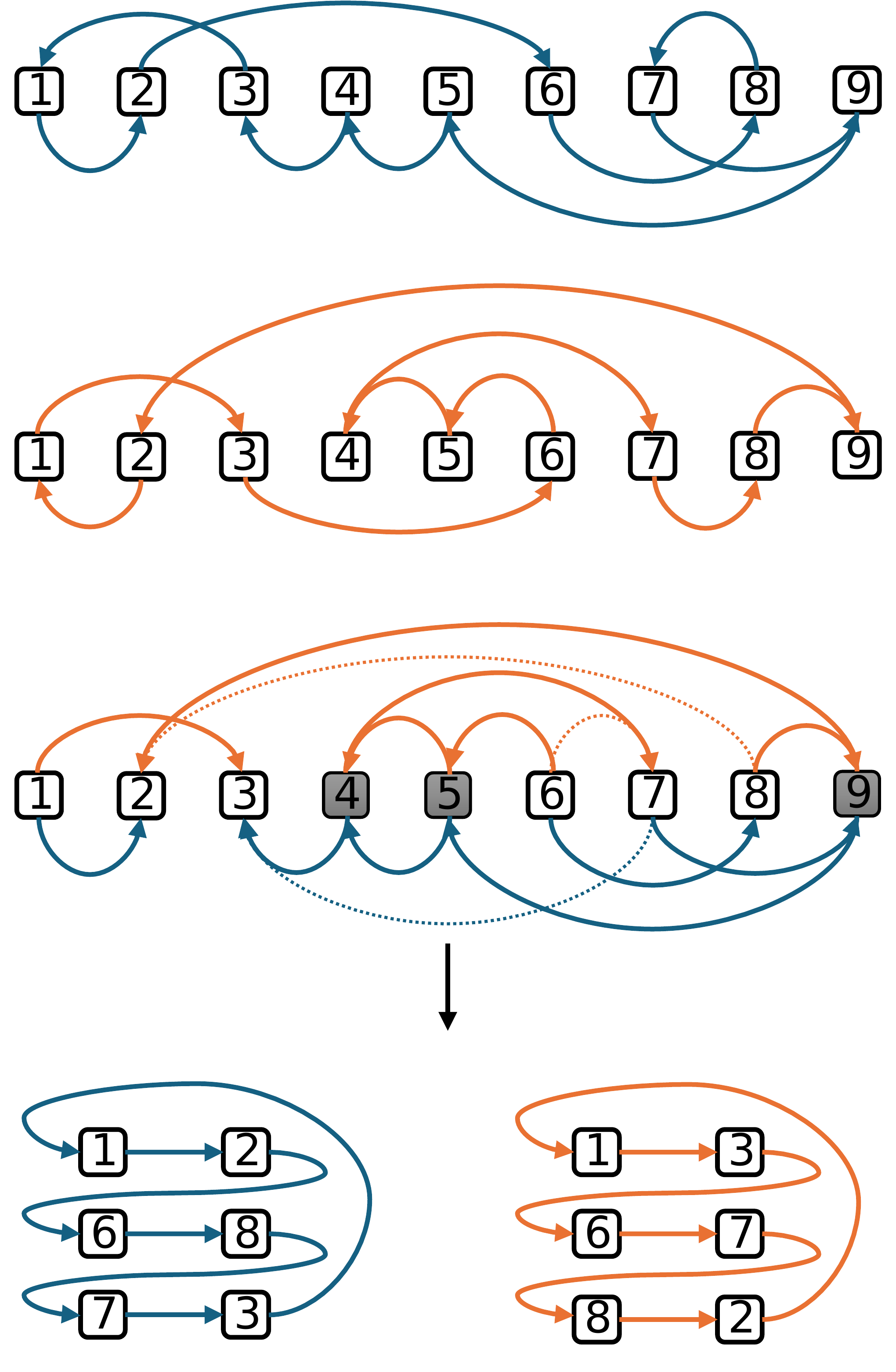}
    \caption{
    The top two images show parent solutions $A$ (blue) and $B$ (orange), respectively. An exemplary AB-cycle of these two parents, with $B$-vertices marked in gray, portals in white, and the simplified inner connections between portals depicted using dashed lines, is shown in the third image. The final two images illustrate the (four arrays of) inner and outer portals of parent $A$ (left) and parent $B$ (right). Note that the external edges are not part of the AB-cycle and thus not depicted in the exemplary AB-cycle (third row).
    }
    \label{fig:inner_outer_ports}
\end{figure}

After iteratively joining all paths, there are two possible scenarios: (1) If a single path remains, $c_{AB}$ produces a valid Hamiltonian circle, i.e., there are no subtours. (2) If multiple disjoint paths remain, $c_{AB}$ introduces subtours, and the number of subtours equals the number of remaining disjoint paths. 
This two outcomes are trivially necessary and sufficient to determine if an AB-cycle results in one ore more subtours, as this represents the full procedure.
In this process, the vertices that are not endpoints of the paths (i.e., they are $A$- or $B$-vertices) play no role in determining the number of subtours. By the concept of an AB-cycle, it is ensured that all nodes are present twice in the paths of $c_{AB}$: once from parent $A$ and once from parent $B$. Therefore, all points will always be present in the newly produced solution. The endpoints of the paths alone (the $C$-vertices) are sufficient to determine how the paths are joined since only the endpoints determine the connectivity of the paths. The number of subtours created by $c_{AB}$ is identical to the number of subtours that would be created if the paths were simplified to include only their endpoint nodes (the $C$-vertices). $\square$ 

Using this theorem, and given an AB-cycle $c_{AB}$ with vertices $V_c$, the check can be performed as follows. First, identify the set of portals (i.e., $C$-vertices) $V_p \subseteq V_c$. All vertices that are not in $V_c$ are external ($A$-vertices) and, thus, no portals. $B$-vertices are contained in $V_c$, but can be identified as nodes connected with four inner edges, i.e., which appear twice in the chain of vertices of $c_{AB}$. 
Next, we determine inner and outer portals for parents $A$ and $B$. Given a parent solution, inner portals are the ones where the parent enters the AB-cycle and outer portals are those where it leaves the AB-cycle. This yields four arrays -- inner and outer portals for $A$ and $B$, respectively -- with $n = |V_p| / 2$ portals. 
If they are sorted by the order of appearance in the respective tour, internal and external links are easy to follow, see \cref{fig:inner_outer_ports}.

Let $P_{I,A} := (p_{I,A,1}, \dots, p_{I,A,n})$ and $P_{O,A} := (p_{O,A,1}, \dots, p_{O,A,n})$ be the sorted arrays of internal and external portals, respectively, for parent $A$. If the first portal appearing in $A$ is an inner portal, the $i$-th inner connection is $(p_{I,A,i},p_{O,A,i})$ and the $i$-th outer connection is $(p_{O,A,i},p_{I,A,i + 1})$. 
If the first portal appearing in $A$ is an outer portal, the $i$-th inner connection is $(p_{O,A,i},p_{I,A,i + 1})$ and the $i$-th outer connection is $(p_{I,A,i},p_{O,A,i})$. 
Starting from the first inner portal of $B$, cycles can be easily found by alternately tracing of inner connections of $B$ and outer connections of $A$.
As the arrays are sorted, searching for a corresponding portal of the other parent is efficient even for larger AB-cycles, as every portal has only one inner connection from parent $B$ and one outer connection from parent $A$. Thus, the first portal that is visited twice must be the starting portal. Checking if it is reached with the current connection is sufficient to encounter a subtour, if not all of its portals are visited. Once such a subtour is found, and if the exact number of subtours is of interest, the procedure can be restarted with one of the remaining portals that were not yet visited. The number of times that a new portal must be chosen equals the number of subtours.

\section{Benchmarking New EAX-Variants}
\label{sec:variants}

\begin{table}[t]
    \centering
    \caption{Average time (in seconds) per instance, differentiated by the four set sizes, for checking whether an AB-cycle will break parent $A$ or not and for repairing an invalid solution.}
    \label{tab:rep-check-time}
    \begin{tabular}{rcccc}
        \toprule
                    & \multicolumn{4}{c}{Instance Size} \\
        Time for       & $500$ & $1\,000$ & $1\,500$ & $2\,000$ \\
        \midrule
        Checking    & 2.2 & \phantom{1}6.5 & 10.2 & 14.4 \\
        Repairing   & 3.9 & 16.5 & 32.0 & 62.2 \\
        \bottomrule
    \end{tabular}
\end{table}

Based on the observations from \cref{sec:examination}, two promising EAX variants can be derived, both leveraging the fast check for complete tours, introduced in \cref{sec:check}. As summarized in \cref{tab:rep-check-time}, determining whether an AB-cycle produces a valid solution is computationally less expensive than repairing a broken tour, particularly for larger problem instances . Furthermore, AB-cycles that do not require repair are frequently favored by the selection criterion (\cref{fig:heatmap}) and are effective in introducing optimal edges into the new solution (\cref{fig:heatmap_ratios}). Therefore, the first EAX-variant proposed in here (only-complete), restricts stage I to AB-cycles that result in a valid solution without requiring repair. 
However, as shown in \cref{fig:heatmap_ratios}, the repair process can also introduce new optimal edges that may no longer be present in the current population. To account for this, the second proposed EAX variant (ratio-based) permits all AB-cycles, where the ratio of introduced optimal edges to deleted optimal edges on average exceeds one. This selection excludes AB-cycles associated with the reddish tiles in \cref{fig:heatmap_ratios}, ensuring that only beneficial cycles are utilized.

Since eliminating non-beneficial AB-cycles increases the selection frequency of the remaining ones, it may lead to an excessive number of sampled AB-cycles, reducing efficiency. To address this, we experiment with different settings for the number of offspring generated per parent pair. Our findings indicate that, for the evaluated set of instances, an optimal setting is 20 offspring, which we adopt as the fixed parameter for stage II.

\begin{figure}[t]
    \centering
    \includegraphics[width=\columnwidth]{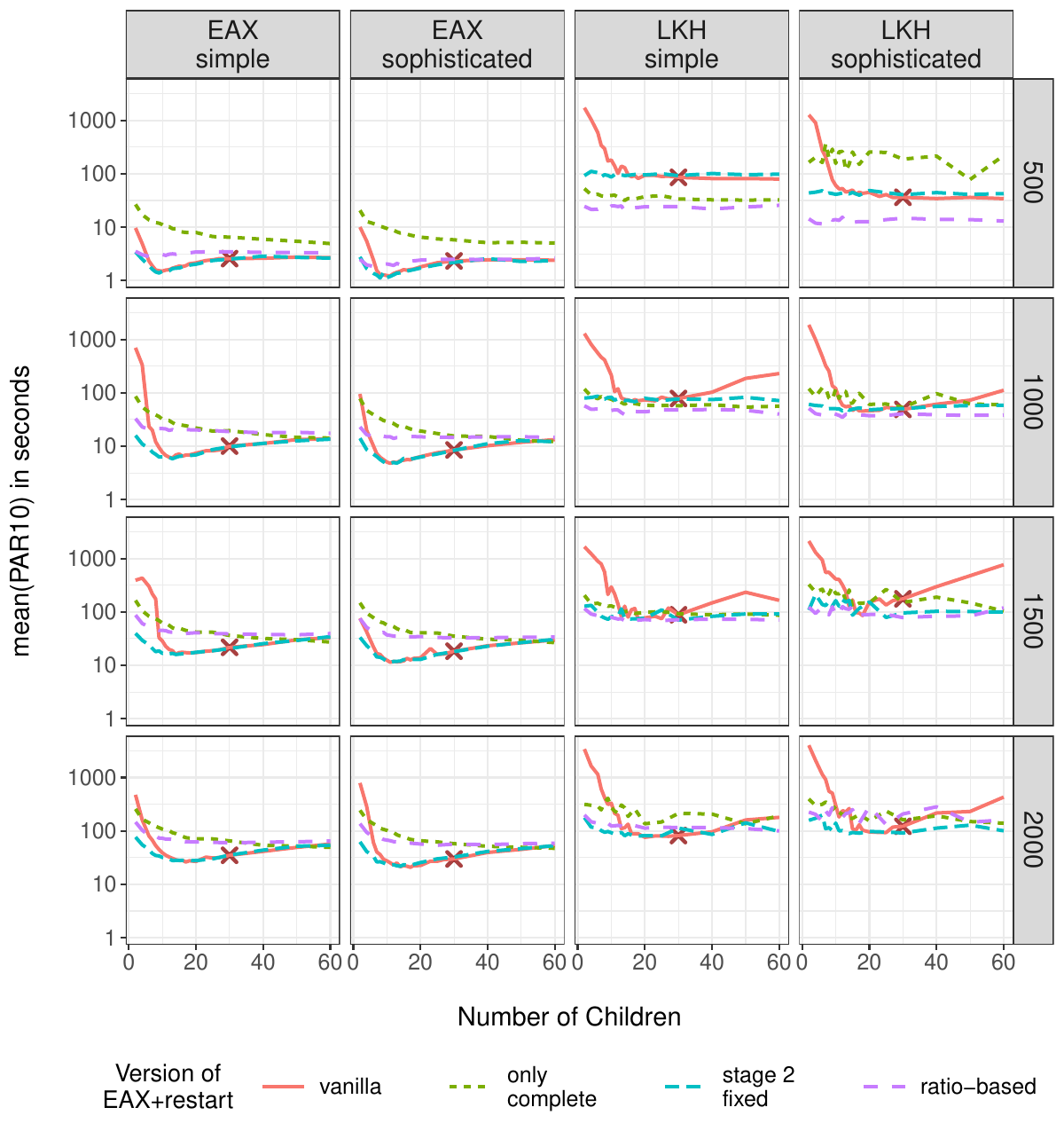}
    \caption{Comparing the performances (averaged PAR10 scores) of four EAX+restart variants: the vanilla-version (solid red curve), a version that only uses AB-cycles resulting in complete tours (green), a version in which the number of children is fixed for stage II (blue), and the ratio-based variant proposed in this work (violet). The results are shown for different types and sizes of TSP sets and across varying numbers of produced children. The default of EAX+restart uses 30 children, whose performance is marked with a red cross. Notably, on the small LKH friendly instance the new ratio-based version is able to solve the problems in less then half the time of the vanilla restart version which is highly influenced by the number of children parameter.} 
    \label{fig:perf_curves}
\end{figure}

\Cref{fig:perf_curves} presents the performance results across different algorithm configurations, each run $10$ times on $100$ instances of the respective group. The performance is measured in terms of the penalized average runtime (PAR10), which penalizes timeouts by setting the respective runtime to ten times the cutoff time (one hour in our experiments), and subsequently computes the mean over all runs per instance. First, we examine the vanilla-version of EAX+Restart (red curve), i.e., the current state of the art, while varying the number of offspring simultaneously in both stage I and stage II. Next, we compare this to a version, where the number of offspring is adjusted only in stage I (blue). Finally, we evaluate our two proposed configurations, i.e., only considering complete tours (green) and the ratio-based variant (purple). Both variants also use a fixed number of offspring in stage II. 

Interestingly, the default configuration of (vanilla) EAX+restart, which uses 30 offspring, consistently proves excessive, leading to suboptimal performance. Moreover, for certain instance types, an increased number of sampled candidate AB-cycles appears to hinder the search for the optimal solution, beyond the additional computational cost of generating more candidates. Fixing the number of offspring in stage II significantly reduces the performance variance observed in the vanilla EAX version. This suggests that stage I has a limited impact on overall performance and that improvements achieved by modifying stage I alone are more substantial.

Our new variants perform worse than the vanilla EAX on EAX-easy instances. However, these instances were constructed to be easily solvable by this particular version of EAX \cite{bossek2019}. As a result, they may be structured in a way that mitigates the negative effects of less useful AB-cycles within the overall population.
In contrast, the two new variants, particularly the ratio-based version (purple curve), appear to address challenges in evolved instances, which were challenging for the original approach. Notably, particularly for smaller instances, the average PAR10 score is significantly reduced, with, on average, less than 50\% of the runtime (needed for solving an instance) compared to state-of-the-art approaches. For larger instances, stage II of EAX -- which we did not modify in this work -- may have a greater influence, potentially limiting performance gains.
Additionally, the different operators used to develop the simple and sophisticated TSP sets appear to have caused distinct patterns, as reflected in the volatile performance of the only-complete EAX-variant on LKH-friendly instances of size 500.

\begin{figure}[t]
    \centering
    \includegraphics[width=\columnwidth]{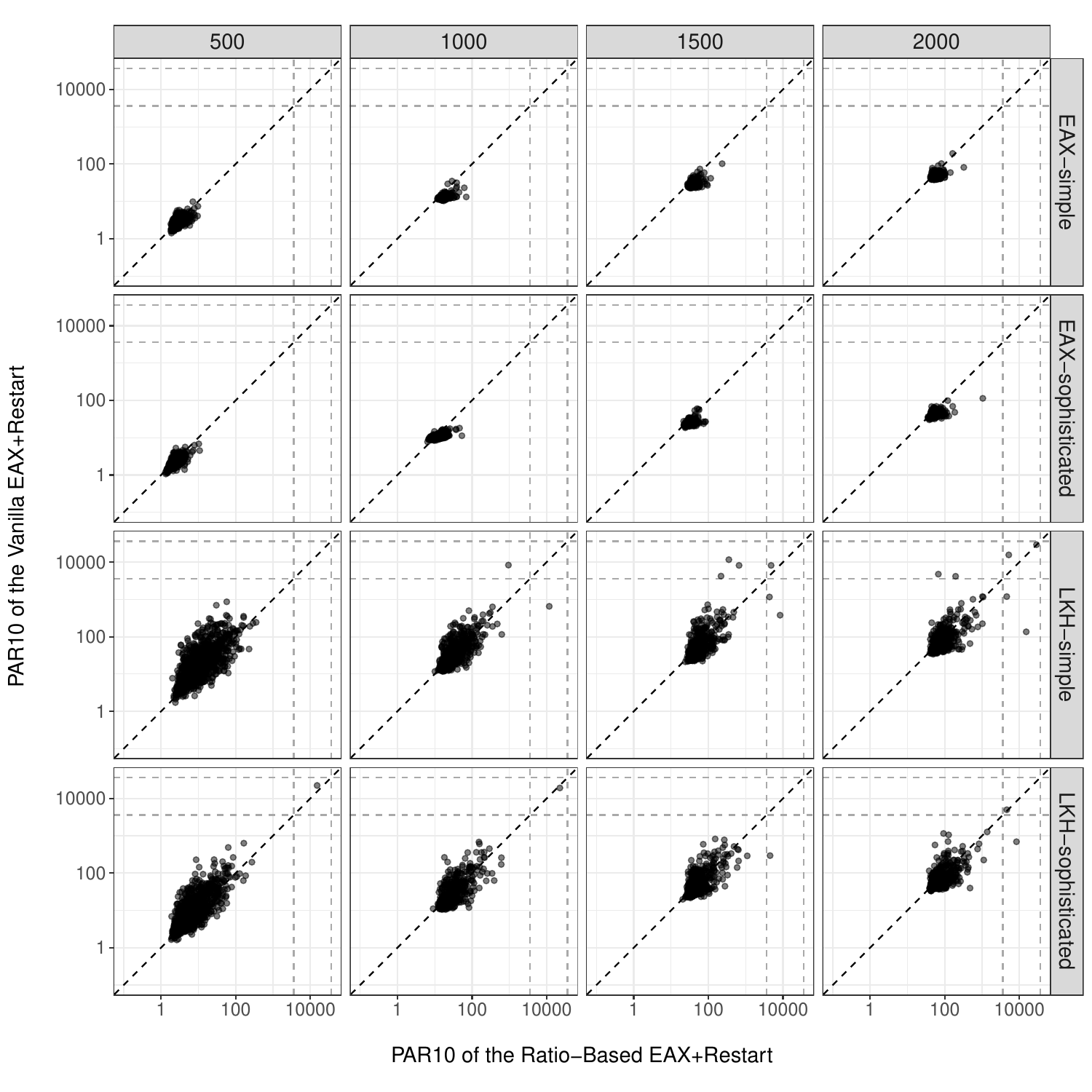}
    \caption{Comparison of the PAR10 scores of our ratio-based EAX+Restart (x-axis) with its vanilla version, representing the current state-of-the-art (y-axis). Our approach is superior to the state-of-the-art on small LKH-friendly instances and performs on par for small EAX-friendly instances. With increasing instance size, both EAX variants perform comparably on LKH-friendly instances, while the vanilla EAX might be slightly better on EAX-friendly instances -- which were tailored to this particular EAX variant.}
    \label{fig:eax_scatter}
\end{figure}

At last, we directly compare the performances of the vanilla and the ratio-based versions of EAX+restart in \cref{fig:eax_scatter} on all instances. We find that our ratio-based version improves runtime on many of the LKH-friendly instances, especially of size 500, rather than just avoiding a few outliers. For other instance types, i.e., large LKH-friendly instances, as well as the EAX-friendly instances, the performance remains mostly comparable to the vanilla version.

\section{Conclusion}
\label{sec:conclusion}

This paper investigates the first stage of EAX, the current state-of-the-art heuristic for solving the TSP. In particular, we examine how different types of AB-cycles will evolve during the algorithm's evolutionary loop, and how they affect the solver's performance.
Building on our findings, we introduce a novel check -- for which we show that it is both necessary and sufficient -- to quickly determine whether an AB-cycle results in a valid tour, or whether it splits the tour into subtours. In case of the latter, it also determines how many subtours are created. Beyond its relevance to this study, particularly in the context of stage I of the EAX algorithm, this check has broader implications for future research. For instance, it opens avenues for refining stage II of EAX, as well as enhancing powerful crossover operators such as GPX by providing more valid recombining components. 

The analysis of different AB-cycle types -- based on the number of optimal edges that were introduced or removed -- revealed substantial potential for performance improvements on smaller LKH-friendly instances. For instance, the ratio-based variant of EAX proposed within this work, demonstrated superior performance on LKH-friendly instances of size 500, achieving performances that halved the average computational time compared to the state-of-the-art restart-version of EAX. At the same time, it achieved comparable performances across most of the remaining TSP sets.
Hence, our results emphasize the potential for novel findings and advancements in both the theoretical understanding and practical application of AB-cycle-based recombination algorithms. They also hint at a potential source of structural differences between instances favorable to LKH and those favoring EAX, which merit further investigation in future studies.

Importantly, our studies also reveal that limiting the number of offspring produced by EAX enhances its overall performance. This suggests that (a) EAX itself needs to be further studied to deepen our understanding of its strengths and weaknesses, and (b) there is a high chance that automated algorithm configurators will find a more powerful default configuration of EAX.

\begin{acks}
The authors gratefully acknowledge the computing time made available to them on the high-performance computer at the NHR Center of TU Dresden. This center is jointly supported by the Federal Ministry of Education and Research and the state governments participating in the NHR (\url{www.nhr-verein.de/unsere-partner}).
This research received financial support from the German Academic Exchange Service (DAAD) under the Program for Project-Related Personal Exchange (PPP).
Darrell Whitley was supported by US National Science Foundation Grants, Award Number 1908866 and CCF Award Number (FAIN) 2426840.
Pascal Kerschke acknowledges support by the \href{https://scads.ai}{\em Center for Scalable Data Analytics and Artificial Intelligence (ScaDS.AI) Dresden/Leipzig}.
\end{acks}

\bibliographystyle{ACM-Reference-Format}
\bibliography{sample-base}










\end{document}